\documentclass[12pt]{spieman}  
\usepackage{amsmath,amsfonts,amssymb}
\usepackage{graphicx}
\usepackage{setspace}
\usepackage{tocloft}
\usepackage[titlenumbered,ruled]{algorithm2e}
\usepackage[normalem]{ulem}
\usepackage{color,soul}
\title{Discriminative Dictionary Design for Action Classification in Still Images and Videos}

\author[a]{Abhinaba Roy}
\author[b]{Biplab Banerjee}
\author[c]{Amir Hussain}
\author[d]{Soujanya Poria}
\affil[a]{Nanyang Technological University, Singapore}
\affil[b]{Indian Institute of Technology, Bombay, India}
\affil[c]{Edinburgh Napier University, UK}
\affil[d]{Singapore University of Technology and Design, Singapore}

\cftpagenumbersoff{figure}
\cftpagenumbersoff{table} 
\begin{document} 
\maketitle

\begin{abstract}
In this paper, we address the problem of action recognition from still images and videos. Traditional local features such as SIFT, STIP etc. invariably pose two potential problems: 1) they are not evenly distributed in different entities of a given category and 2) many of such features are not exclusive of the visual concept the entities represent.
In order to generate a dictionary taking the aforementioned issues into account, we propose a novel discriminative method for identifying robust and category specific local features which maximize the class separability to a greater extent.
Specifically, we pose the selection of potent local descriptors as filtering based feature selection problem which ranks the local features per category based on a novel measure of distinctiveness. The underlying visual entities are subsequently represented based on the learned dictionary and this stage is followed by action classification using the random forest model followed by label propagation refinement. The framework is validated on the action recognition datasets based on still images (Stanford-40) as well as videos (UCF-50) and exhibits superior performances than the representative methods from the literature.  
\end{abstract}

\keywords{Action recognition, Local features, Feature mining, Random forest}


\begin{spacing}{2}   

\section{Introduction}
\label{sect:intro}  
Recognition of visual concepts is one of the most active research areas in computer vision. Especially human action recognition from images and videos have been popular amongst researchers in recent times. With growing amount of visual data available from various sources, intelligent analysis of human attributes and activities has gradually attracted the interest of the computer vision community. One of the widely used approaches in action recognition is based on local descriptors that are based on three stages: 1) Extraction of local descriptors, 2) Codebook (dictionary) generation and feature encoding, and 3) Classification based on the encoded features. Efficiency of such a model depends upon a number of factors, and effective codebook generation is undoubtedly the most noteworthy.

Standard codebook generation process is based on vector quantization of local descriptors extracted from the available training data in which the cluster centroids define the codewords; the basic building blocks that are ultimately used to encode the underlying visual entities. Specifically, an entity is represented by a vector where the $i^{th}$ component can be either the number of local descriptors that fall in the $i^{th}$ cluster or a measure of proximity of local descriptors to the $i^{th}$ cluster centroid.
Needless to mention, the quality of the extracted local descriptors affect representation power of the codewords which, in turn, has direct impact on the recognition performance. For instance, the descriptors extracted from background regions or the ones shared by many visual categories add little to the discriminative capability of the codebook in comparison to the ones specifically extracted from the objects of interest. 
However, it is impossible to ensure the selection of potentially useful local descriptors in advance since such feature extraction techniques are typically engineered and ad hoc. In other words, there are certain immediate advantages if the most discriminative local descriptors are used for the purpose of a cogent codebook construction, though the process is intrinsically complex in general. Selection of the discriminative local descriptors for effective codebook generation coping with action recognition from images is the very core topic of this paper. We propose a simple algorithm which gradually filters out unrepresentative descriptors before constructing a compact global codebook. The proposed method is generic in the sense that it can work with different types of local features irrespective of the underlying visual entities they refer to. Specifically, we represent each still image by a large pool of category independent region proposals \cite{alexe2012measuring}. Each region proposal is represented by convolutional neural network (CNN) features ($4096$ dimensions) obtained from a pre-trained network.
We propose a sequential method for codebook construction which first clusters the local descriptors of each entity using the non-parametric mean-shift (MS) technique  \cite{comaniciu2002mean}. The cluster centroids thus obtained represent the reduced set of non-repetitive local features for the entity from now onwards. Another round of MS clustering on the new set of local descriptors calculated from all the entities of a given category is followed and the centroids thus obtained are employed to build a temporary codebook specific to each category.
Further, we propose an adaptive ranking criteria to highlight potentially discriminative codewords from each category specific codebook and the global dictionary is built by accumulating these reduced set of codewords from all the categories. Efficient codebook construction is not new to the computer vision literature. However, we argue that our codebook construction technique explicitly incorporate the class support and a novel notion of distinctiveness based on conditional entropy is introduced. 

We can summarize the main highlights of this paper as follows:

\begin{itemize}
\item The initial two level MS based clustering of the local descriptors on the entity and the category level largely reduces the effects of repetitive and uninteresting descriptors, yet selecting representative codewords from each locally dense region in the feature space. We further propose a novel adaptive measure to rank and select a subset of discriminative codewords per visual category under consideration using the concepts of conditional entropy and term frequency- inverse document frequency (tf-idf) score followed by an adaptive ranking technique. The proposed ranking method ensures that the selected set of codewords are frequent in the entities of the same category while being sporadic in other visual categories.
\item We evaluate the codebooks learned in this way for action categorization from still images. We observe that the learned codebooks, when used in conjunction with efficient feature encoding techniques, sharply outperform similar techniques from the literature. Specifically, considering the size of the local descriptors, we consider the locality constrained linear coding (LLC) \cite{wang2010locality} for action recognition in images and Fisher vector \cite{perronnin2010improving} for videos. Consecutively, classification is performed using random forest classifier.
\item As post processing of classification results, we apply label propagation algorithm to improve the classification of random forest.

\end{itemize}

The rest of the paper is organized as follows. We discuss a number of related works from the literature in Section \S \ref{sec2}. The proposed action recognition framework is described in Section 3. Experimental results are reported in Section 4, followed by concluding remarks and ideas of possible future endeavour.

\section{Related Work}
\label{sec2}
In this section, we highlight two aspects of the proposed framework and discuss relevant techniques from the literature. First, action representation from images and videos with a focus on local feature encoding based methods is addressed and  a discussion on the relevant codebook construction techniques is subsequently be followed. 
\subsection{Action recognition from still images and videos - use of local features}
Recognition of human actions and attributes \cite{cheng2015advances} has been approached using traditional image classification methods \cite{yang2010recognizing}. In the standard dictionary learning based scenario, a typical framework extracts dense SIFT \cite{lowe2004distinctive} from the training images and codebook is constructed by clustering the SIFT descriptors by k-means clustering. Further, efficient encoding techniques including bag of words (BoW), LLC, Fisher vector are used to represent the images before the classification stage is carried out in such a feature space \cite{chatfield2011devil}. 
Since the inherent idea of the BoW based frameworks is to learn recurring local patches, a different set of approaches directly models such object parts in images. Such techniques either initially define a template and try to fit it to object parts or iteratively learn distinctive parts for a given category.
Discriminative part based models (DPM) \cite{felzenszwalb2010object} are used extensively for this purpose and they served as the state of the art for a period. 
The hierarchical DPM model is used to parse human pose for action recognition in \cite{wang2012discriminative}. An efficient action and attributed representation based on sparse bases of local features is introduced in \cite{yao2011human}. An expanded part model for human attribute and action recognition is proposed in \cite{sharma2013expanded}. The effects of empty cavity, ambiguity and pooling strategies are explored in order to design the optimal feature encoding for the purpose of human action recognition in still images in \cite{zhang2016towards}. 
Very recently, the part learning paradigm has gained much attention because of its ability to represent mid-level visual features. Given a large pool of region proposals extracted from the images, such techniques iteratively learn part classifiers with high discriminative capabilities. Methods based on partness analysis \cite{juneja2013blocks}, deterministic annealing for part learning \cite{sicre2015discriminative} etc. are some of the representatives 
in this respect. The notion of a part is further extended to videos by constructing spatio-temporal graphs of the local keypoints over the video frames \cite{zhou2015interaction}.
Similarly, the BoW framework exhibits impressive performance in recognizing action from video data.
Broadly, the video level features can be categorized into hand-engineered and deep features. The popular hand-crafted feature set includes STIP \cite{laptev2005space}, selective STIP \cite{chakraborty2011selective}, dense and improved trajectory \cite{wang2013action} and optical flow based features \cite{xiong1998efficient}. Further, several descriptors are used to encode the scene around such detected keypoints (HOG, HOF etc.). In contrary, the deep CNN structures for videos combine separate models for static frames and the inter-frame motion. While a standard image based CNN models (AlexNet, GoogleNet, VGGNet etc.) can be used to extract per frame features, sophisticated optical-flow CNN is modeled for capturing the motion efficiently \cite{simonyan2014two}. 

\subsection{Dictionary learning}
Initial works in dictionary learning proposes k-means clustering to create a dictionary, followed by a Bag of Words (BoW) encoding \cite{shukla2013action,bettadapura2013augmenting}. Amongst alternate approaches that focus on Sparse coding, Qiu et al. \cite{qiu2011sparse} reports a sparse dictionary-based representation for action. Liu et al. \cite{liu2015hessian} proposes a Hessian regularized sparse coding method for action recognition. Lu et al. \cite{lu2013abnormal} proposed the idea of slicing frame to patches in different scales and using patches to train dictionaries. Fanello et al. \cite{fanello2013keep} introduces a real-time action recognition method with dictionary learning. Xu et al. \cite{xu2017two} proposes a two stream dictionary learning architecture that consists of interest patch (IP) detector and descriptor. 

There are not many approaches that focus on dictionary learning for the task of action recognition. Rather, they are more interested in development of proper feature representations. Contrary to these approaches, in this work we focus on construction of discriminative dictionary and prove the effectiveness  using generic feature representations.
\section{Proposed algorithm}
We detail the proposed action recognition framework in this section. As already mentioned, the proposed framework consists of four major stages:
1) Extraction of local features, 2) Discriminative dictionary construction, 3) Feature encoding, 4) Action classification.

For notational convenience, let us consider that $TR=\{X_i,Y_i\}_{i=1}^N$ constitutes $N$ training examples belonging to $L$ action categories where each $X_i$ represents an image or a video and $Y_i$ is the corresponding class label.  Entities in $TR$ are represented by a set of local descriptors $F_i=\{F_i^1,F_i^2,\ldots,F_i^{\alpha_i}\}$ 
where $F_i^k \in \mathbb{R}^d$ and $d=4096$ or $d=162$, respectively, depending on whether the underlying $X_i$ is an image or a video. In addition, $\alpha_i$ represents the number of local descriptors extracted from $X_i$. Further, $\{C_1,C_2,\ldots,C_L\}$ represents the set of category specific codebooks learned by the proposed algorithm by exploiting the local features extracted from $TR$, whereas $C=[C_1 C_2 \ldots C_L]$ is the global codebook obtained by the concatenation of the local ones.

The framework is elaborated in the following sections.
\vspace{-1em}
\subsection{Extraction of local features}

We consider category independent region proposals to highlight local regions in still images whereas the  popular STIP features are used for video streams.

Region proposal generation techniques highlight region segments in the image where the likelihood of the presence of an object part is high. This provides a structured way to identify interesting locations in the image and thus reduces the search space for efficient codeword generation.
We specifically work with the objectness paradigm for region proposals generation from still images which is based on modeling several aspects regarding the characteristics of the objects in a Bayesian framework. Each region proposal is further represented by the CNN features. We prefer the ImageNet pre-trained VGG-F  \cite{chatfield2014return} model which has an architecture
similar to AlexNet \cite{krizhevsky2012imagenet}, and comprises of 5 convolutional layers and 3 fully-connected layers.
The main difference of VGG-F and AlexNet is that VGG-F contains less convolutional layers and uses a stride of 4 pixels leading to better evaluation speed than the AlexNet architecture.
In case of videos, the representation of local variations depends on local STIP keypoints. STIP features are the extension of the Harris corner detectors for images to the spatio-temporal domain. They are detected at locations where the video frame level intensities have significant local variations in both space and time. Histogram of oriented gradients (HOG) and histogram of optical flow (HOF) features are extracted around each STIP point.

\subsection{Discriminative dictionary learning}

We first build category specific codebooks and then concatenate all the local codebooks to generate a global codebook.

\subsubsection{Separate dictionary learning for each category}
\label{dic}
 For a given $l \in \{1,2,\ldots,L\}$, the dictionary learning process is summarized as follows:

\begin{enumerate}
\item For each training instance with the category label $l$, we first group the local descriptors using MS clustering and consider the cluster 
centroids as constituting the reduced set of local descriptors. MS is an iterative, non-parametric clustering method which does not require an estimation 
of the number of clusters as input. Instead, it relies on the kernel density estimate in the feature space to group samples which form dense clusters. 
Given $F_i=\{F_i^1,F_i^2,\ldots,F_i^{\alpha_i}\}$, the kernel density estimate at a point $F_i^k$ is expressed as

\begin{equation}
f(F_i^k)=\frac{1}{\alpha_i h^d} \underset{m=1}{\overset{\alpha_i}{\sum}} K(\frac{F_i^k-F_i^m}{h})
\end{equation}

where $K$ is a radially symmetric kernel function and $h$ defines the width of the Parzen window to highlight the neighbourhood around $F_i^k$. 
A cluster is identified as the region where the data density is locally maximum. This can alternatively be interpreted as the local regions where
$\nabla f \approx 0$. $\nabla f$ can efficiently be calculated by iteratively shifting the centroids of the Parzen windows until the locally dense regions are reached \cite{comaniciu2002mean}.

Since all the descriptors in a dense region in the feature space highlight near similar local features, the mean-shift clustering is able to select one unique 
representative for all of them. Further, since mean-shift implicitly estimates the number of clusters present in the dataset, hence, the problem of over-merging is greatly reduced. 
On the other hand, spherical clustering techniques like k-means and fuzzy c-means create suboptimal codebooks as most of the cluster centroids fall near 
high density regions, thus under-representing equally discriminant low-to-medium density regions. MS resolves such problem by focusing on locally dense regions in the feature space.
 Let $\widehat{F}_i=\{\widehat{F}_i^1,\widehat{F}_i^2,\ldots,\widehat{F}_i^{\widehat{\alpha_i}}\}$ represents the new set of local descriptors for the $i^{th}$ training instance where each $\widehat{F}_i^k$ represents a cluster centroid.

\item Once $\widehat{F}_i$s are constructed for all the training instances with category label $l$, we vector quantize all such $\widehat{F}_i$s using MS clustering to build a temporary 
codebook $C_l=\{C_l^1,C_l^2,\ldots,C_l^{\beta_l}\}$ for the category with each $C_l^k$ representing a codeword (cluster centroid). 
Similar to the previous stage, it is guaranteed that $C_l$ is ensured to capture all the potential local features for the $l^{th}$ category. 
\end{enumerate}

$\{C_1,C_2,\ldots,C_L\}$ are constructed in the similar fashion for $l \in \{1,2,\ldots,L\}$. It is to be noted that the labels of the codewords depend upon the 
action categories they refer to. Further, the sizes of the $C_l$s may differ from each other. The $C_l$s thus obtained are not optimal in the sense that they contain many codewords with low discriminative property. Such codewords need to be eliminated in order to build robust category specific codebooks.
However, we need a measure to rank the descriptors based on their discriminative ability.
In this respect, the following observations can be made:

\begin{itemize}
\item A potentially discriminative codeword is not frequent over many of the categories constituting the dataset.
\item Most of its nearest neighbors in $\{C_1,C_2,\ldots,C_L\}$ share the same class label with the codeword under consideration. 
\end{itemize}

We model the first observation in terms of the idea of conditional entropy whereas the second observation is replicated by  the tf-idf score.

For a given codeword $C_l^k$, we find out the labels of its $T$ nearest neighbours over the entire set of codewords in $\{C_1,C_2,\ldots,C_L\}$ and subsequently define the conditional entropy measure as:

\begin{equation}
H(Y|C_l^k)= - \underset{l'=1}{\overset{L}{\sum}} p(l'|C_l^k) \log_2 p(l'|C_l^k)
\end{equation}

where $p(l'|C_l^k)$ represents the fraction of the retrieved codewords  with label $l'$.
For discriminative codewords, i.e. the ones which do not span many categories, $H$ is small whereas the value of $H$ grows with the selection of codewords shared by many categories.

In addition to the $H$ score, we also expect the nearest neighbours to be populated from the same category as of $C_l^k$. In order to impose this constraint, 
we define the tf-idf score for $C_l^k$ as follows:

\begin{equation}
\centering
TI(C_l^k)=\frac{|C_{l'}^{k'}|C_{l'}^{k'}\in knn(C_l^k) \ AND \ l'=l|}{|C_{l'}^{k'}|C_{l'}^{k'}\in knn(C_l^k)|}
\end{equation}

Both the measures are further combined in a convex fashion to define the ranking measure as follows:

\begin{equation}
\centering
Rank(C_l^k)= w_1 \ \frac{1}{H(Y|C_l^k)} + (1-w_1) \ TI(C_l^k)
\end{equation}

We repeat this stage for all the codewords in $\{C_1,C_2,\ldots,C_L\}$ .
As already mentioned, the $Rank(C_l^k)$ has high values for potentially discriminative and category specific codewords. 
\subsubsection{Number of Codeword Selection}
\label{sec:code}
We rank the codewords on the basis of the $Rank$ scores. In order to select the number of optimal codeworks to select, we use an adaptive algorithm. This is in stark contrast to the related work done in \cite{Roy2017}, where top $B$ codewords were chosen in a greedy fashion in order to define the final codebook $\widehat{C}_l$ for category $l$. For this adaptive algorithm, we make a spanning tree of the code words $C_l$. We first create a spanning tree $G$ with $C_l$ as nodes and the edge weights as the difference between the features. Note that the nodes are connected sequencially based on the ranked list. On this spanning tree, we carry out a dominant set clustering \cite{pavan2007dominant}. More specifically, we carry out a binary clustering on the graph to get two subgraphs. We select the subgraph (set of codewords) with higher rank. This results in variable numbers of codewords in each of the classes. Algorithm \ref{algo:1} describes the adaptive number of codeword selection process.
\begin{algorithm}[H]
\SetAlgoLined
\KwIn{$C_l$}
\KwOut{$\widehat{C}_l$}
1. Construct a linear chain graph $G = \{C_l,E\}$ with the extracted parts from a given class as the nodes and the edge weights are defined as Euclidean distance between the corresponding features ($E$). \\
3. Perform dominant set clustering on $G$ to to obtain two sub-graphs;\\
4. Choose the subset ($\widehat{C}_l$) with the parts having higher ranks according to the proposed cost function; \\
\caption{Adaptive number of codewords selection}
\label{algo:1}
\end{algorithm}

\subsubsection{Global dictionary construction}
The local codebooks obtained in the previous stage are concatenated in order to obtain a global codebook $\widehat{C}=[\widehat{C_1}\widehat{C_2}\ldots \widehat{C_L}]$. 

\subsection{Feature encoding using $\widehat{C}$}

We represent each visual entity with respect to $\widehat{C}$ for still images and videos separately. 
We find that LLC based encoding works best while dealing with the CNN features in case of action recognition in still images, whereas Fisher vector outperforms other BoW based encoding methods for video based features. For each entity, we consider all the initially extracted local features for encoding.

\subsection{Classification}

The final classification is performed using random forest ensemble classifier \cite{bishop2006pattern}. The decision tree learning algorithm used is information gain and bootstrap aggregation is 
employed to learn the ensemble model. Thus the forest reduces classifier variance without increasing bias. Random subspace splitting is used for each tree split and we consider $\sqrt{d}$ features for each split given $d$ original feature dimensions. The generalization is performed by applying majority voting on the outcomes of the learned trees.
\subsubsection{Label propagation}
\label{sec:lab}
In order to refine and further strengthen the classification results from random forests, we apply an additional round of label propagation. In it's original form, label propagation is a semi-supervised classification way to propagate labels from labeled sample to unlabeled samples \cite{zhu2002learning}. Based on the idea that samples should have same labels if they are neighbors to each other, label propagation "propagates" labels of labeled samples to unlabeled samples according to the proximity. The more similarly samples are, the more easily propagate labels between them. The similarity of samples is calculated as and exponential of negative distance between them (Equation \ref{eq:eq1}).  
\begin{equation}
w_{ij} = exp(-\frac{d^2_{ij}}{\sigma})
\label{eq:eq1}
\end{equation}
where $d_{ij}$ is the distance between $i^{th}$ and $j^{th}$ sample. 
The degree of difficulty for label propagation is described by probabilistic transition matrix ܶ$T_{ij}$, which is defined as  
\begin{equation}
T_{ij} = P(x_i \rightarrow x_j) = \frac{w_{ij}}{\sum_{k=1}^{l+u} w_{kj}}
\end{equation}
where ܶ$T_{ij}$ is the probability of switching label (inference) from instance $x_i$ to $x_j$. Larger $w_{ij}$ leads to larger ܶ$T_{ij}$ that allows labels to propagate more easily. The process of label propagation continues until labels of all samples tend to be stable. Label propagation defines a label matrix ܻ$Y$ with all the label probabilities of data points $x_i$s. 
In this work, we attempt to propagate the labels of most confident labels onto the other inferences with less confidences. First we detect all the output labels from the random forest classifier which most likely provide true inference for a certain test instance. In the cases where there is ambiguity amongst the tree classifier outcomes, we expect propagation of confident labels will solve this issue and provide better classification performance. In order to measure the ambiguity between individual tree classifiers in the random forest, we device a measurement to determine the \textit{confidence} of inference for each test instances. We define the confidence of the inferred labels in terms of agreement of trees in majority voting of random forest. Hence confidence of inferred $y_i$ from a test instance $x_i$ is denoted by 
\begin{equation}
conf_i = \frac{N_{y_i}}{N}
\end{equation}
where $N_{y_i}$ denotes the number of trees which inferred $y_i$ as the label of $x_i$ and $N$ is the total number of trees. We rank the test samples in order of confidence of inferences from them. Consecutively, we use top $20\%$ confident inferences and apply label propagation to improve the outcome of the rest of the inferences. The algorithm related to label propagation is shown below.   
\begin{algorithm}[H]
\SetAlgoLined
\KwIn{ Test instances with confident inferences $(x_1,y_1),...,(x_l,y_l)$. Other test instances with not confident inferences $(x_{l+1},y_{l+1}),...,(x_{l+u},y_{l+u})$. $y_i$ is the inference from random forest for instance $x_i$. $Y_L = y_{1},...,y_{l}$; $Y_U = y_{l+1},...,y_{l+u}$; $Y = [Y_U;Y_L]$}
\KwOut{Labels of all instances}
1. Calculate $w_{ij}$ between samples; \\
2. Calculate probabilistic transition matrix $T$;\\
3. Structure matrix Y;\\
4. $TY\rightarrow Y$; \\
5. Normalize $Y$; \\
6. Reset $Y_L$;\\
7. Repeat step 4 until $Y$ does not change/maximum number of iteration is reached;
\caption{Label propagation for improved classification}
\end{algorithm}

\section{Experimental details}
\subsection{Dataset}
We consider the Stanford-40 \cite{yao2011human} still image action recognition database and UCF-50 \cite{reddy2013recognizing} video based action recognition dataset to evaluate the effectiveness of the proposed framework. Stanford-40 actions is a database of human actions with $40$ diverse action types, e.g. brushing teeth, reading books, blowing bubbles, etc. The number of images per category ranges between $180$ to $300$ with a total of $9352$ images. We use the suggested \cite{yao2011human} train-test split with $100$ images per category as training and remaining for testing.
On the other hand, the UCF-50 dataset contains videos representing $50$ actions in a unconstrained environments. The dataset contains a total of $6700$ videos with about $100-150$ videos per category. This dataset is a superset of the popular UCF-11 dataset. We randomly select $60 \%$ of the videos per category to represent the training set and the remaining $40 \%$ is used to evaluate the classification performance of the proposed framework.
\subsection{Experimental setup}
The following experimental setup is considered in order to evaluate the performance of the proposed framework for both the datasets. 

\begin{itemize}
 \item MS clustering is used in conjunction with the Gaussian kernel. The adaptive bandwidth parameter ($h$) is fixed empirically as $\frac{D}{m}$ ($1 \leq m \leq 10$), where $D$ is the average pairwise distance of all the local descriptors extracted from all the visual entities of each category. The same setup is repeated for MS clustering in the entity and the category levels(section \S \ref{dic}).
 
 \item We extract $500$ region proposals per image for the Stanford-40 dataset. Figure 1 depicts the extracted region proposals for a pair of images from the dataset. We further discard proposals which are largely overlapping to each other (overlap of $\geq 50 \%$) in order to highlight potentially discriminative local patches in the images. The STIP keypoints are extracted from the videos using the publicly available implementation of \cite{laptev2005space}. 
 \item The number of final distinctive codewords selected for each class are set adaptively as discussed in section \S \ref{sec:code} . As for feature encoding, for LLC, $100-200$ nearest neighbors per local descriptor are considered to encode the images. We select the optimal hyper-parameters by cross-validation.
 Each image in the Stanford-40 dataset is optimally represented by a sparse vector of length $8000 \times 1$ ($100$ neighbors in LLC).Whereas each video has a feature length of $32400 \times 1$ ($2 \times \mathrm{feature \ dimension} (162) \times \mathrm{No. \ of \ Gaussian \ components} (100)$).
 
 \item Each component tree in the random forest model is essentially a classification and regression tree (CART) \cite{bishop2006pattern}. We conduct experiments with random forest of different sizes ($500-2000$) and find that a random forest with $1000$ CART trees exhibits superior performance. 
 \item For post processing based on label propagation, we fix the threshold for "confident" inferences to $20\%$. This value is determined by $5$ fold cross validation for both image and video data. 
 \item We compare the overall classification performance of the proposed technique with the representative techniques from the literature. All the experiments are repeated multiple times and the average performance measures are reported. 
\end{itemize}

\begin{figure*}
 \centering
 \fbox{\includegraphics[height=30mm, width=35mm]{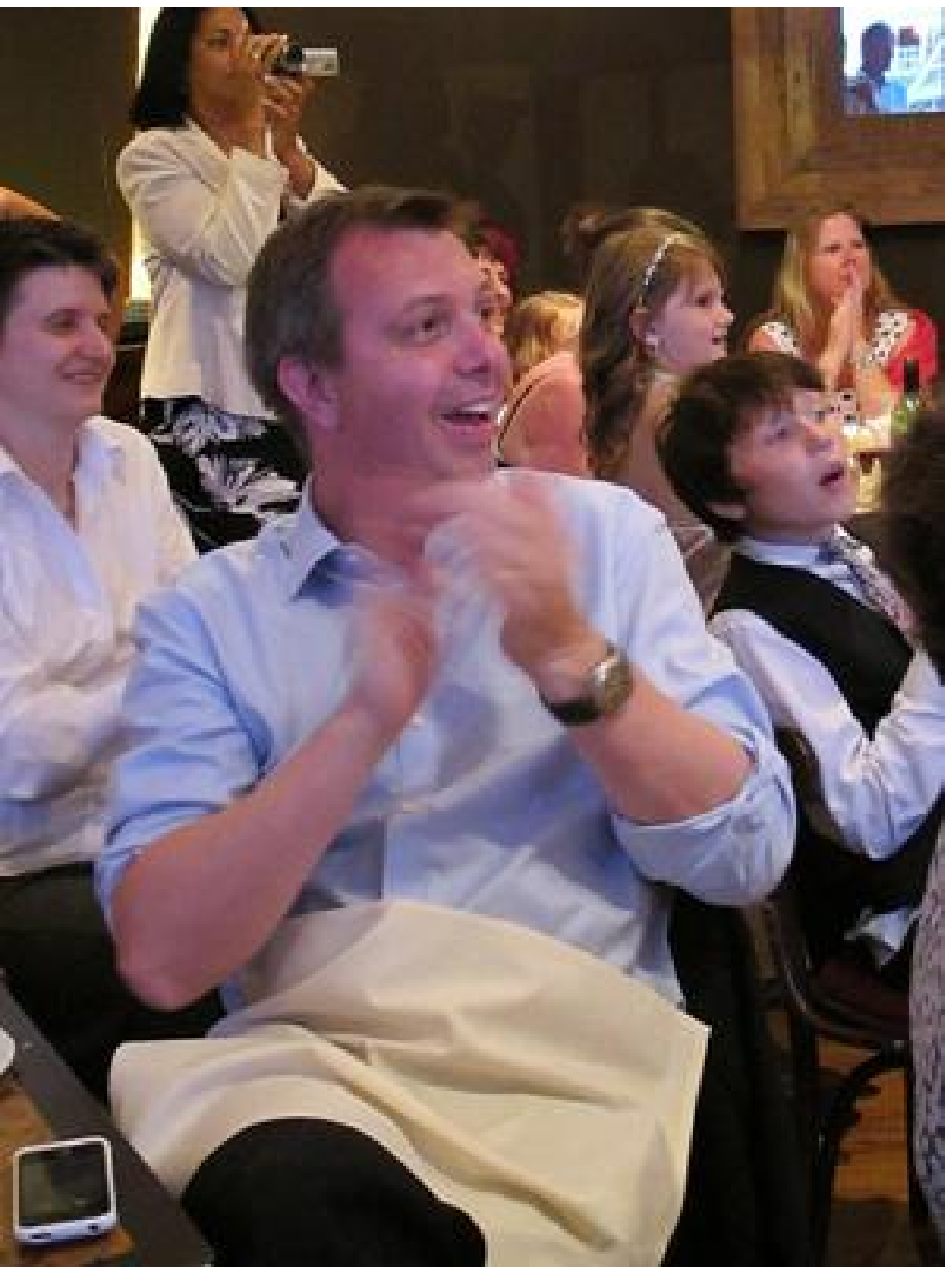}
 \hspace{2px}
 \includegraphics[height=30mm, width=35mm]{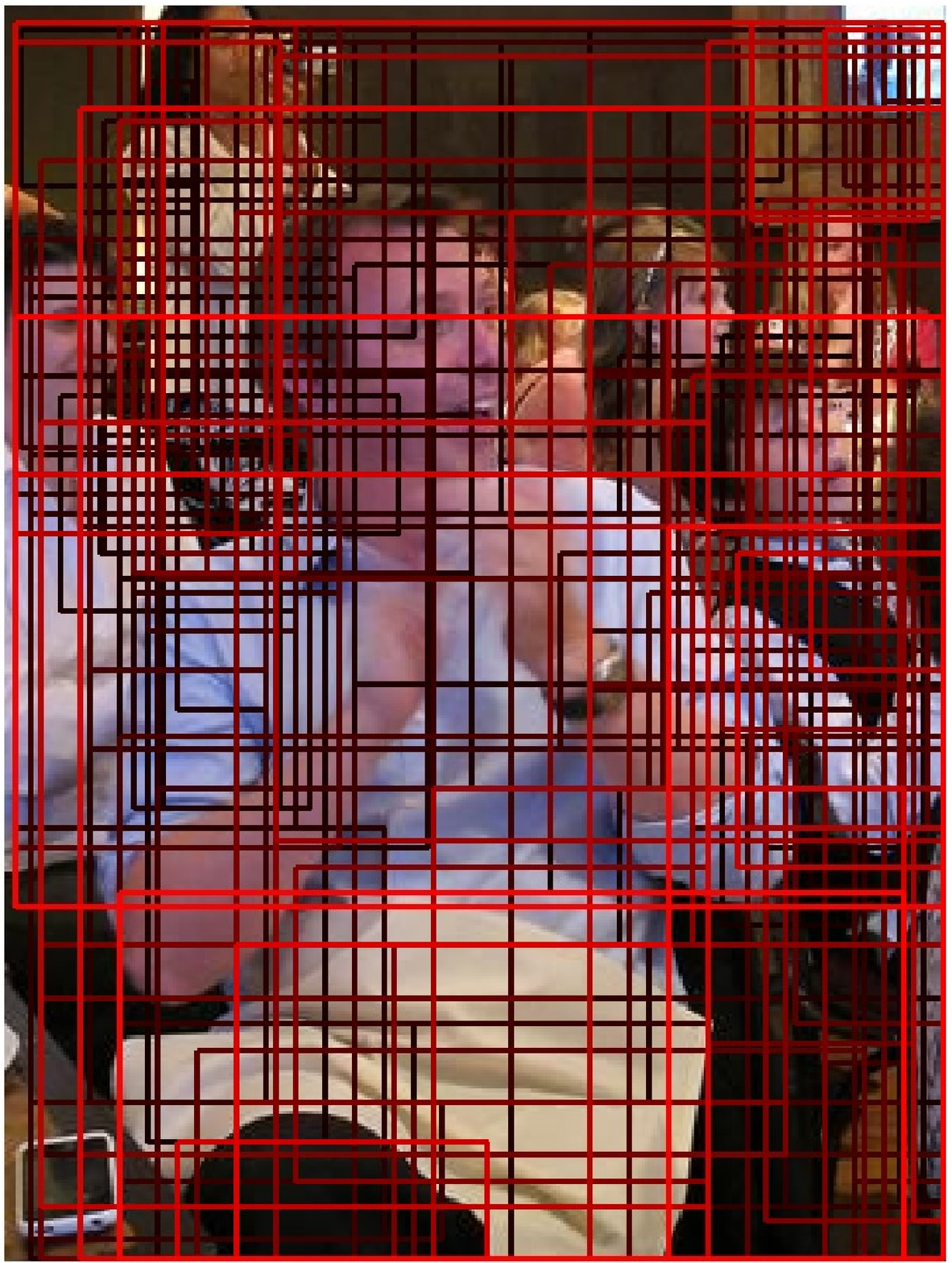}}
 \hspace{10px}
  \fbox{\includegraphics[height=30mm, width=35mm]{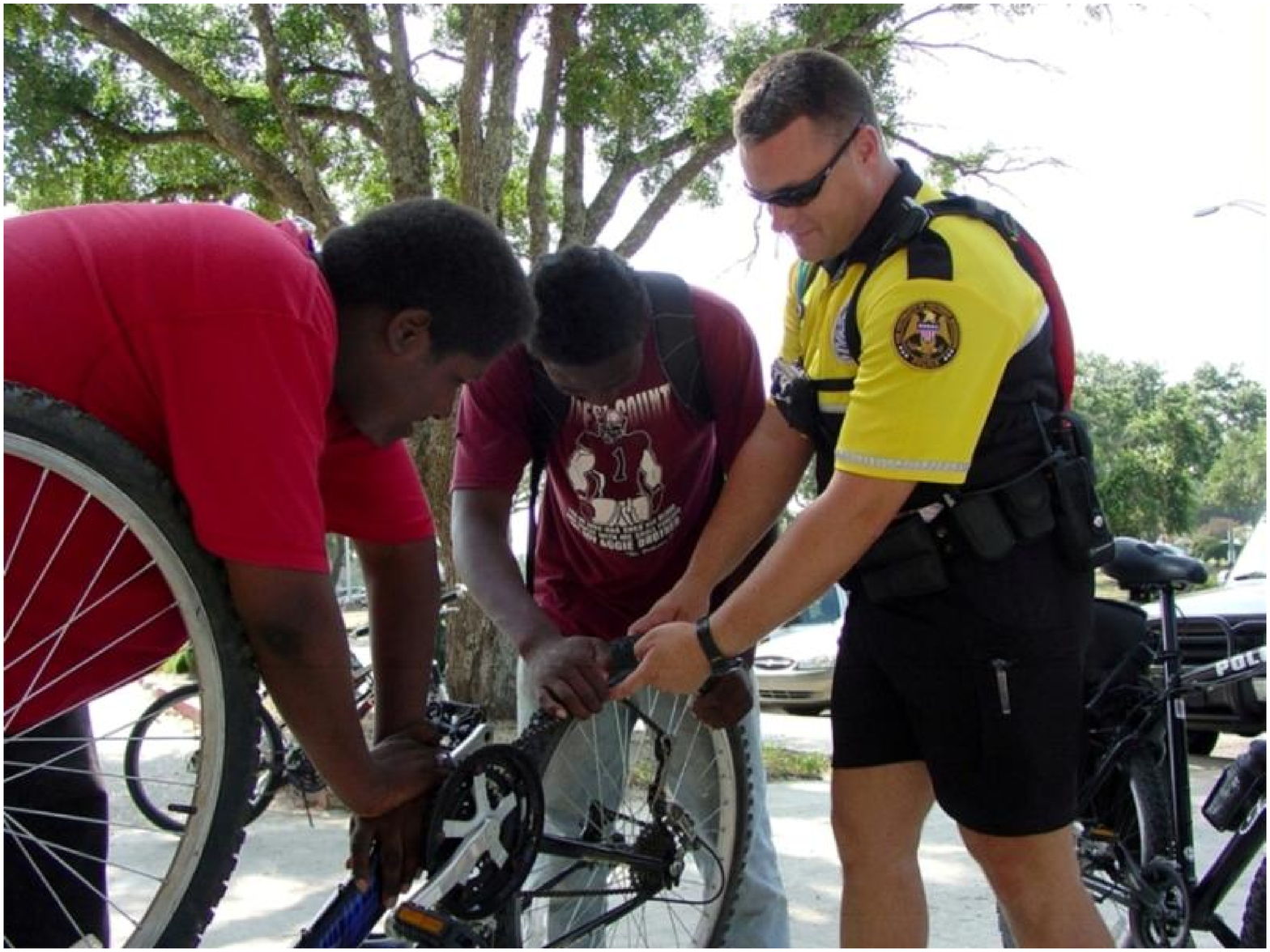}
  \hspace{2px}
    \includegraphics[height=30mm, width=35mm]{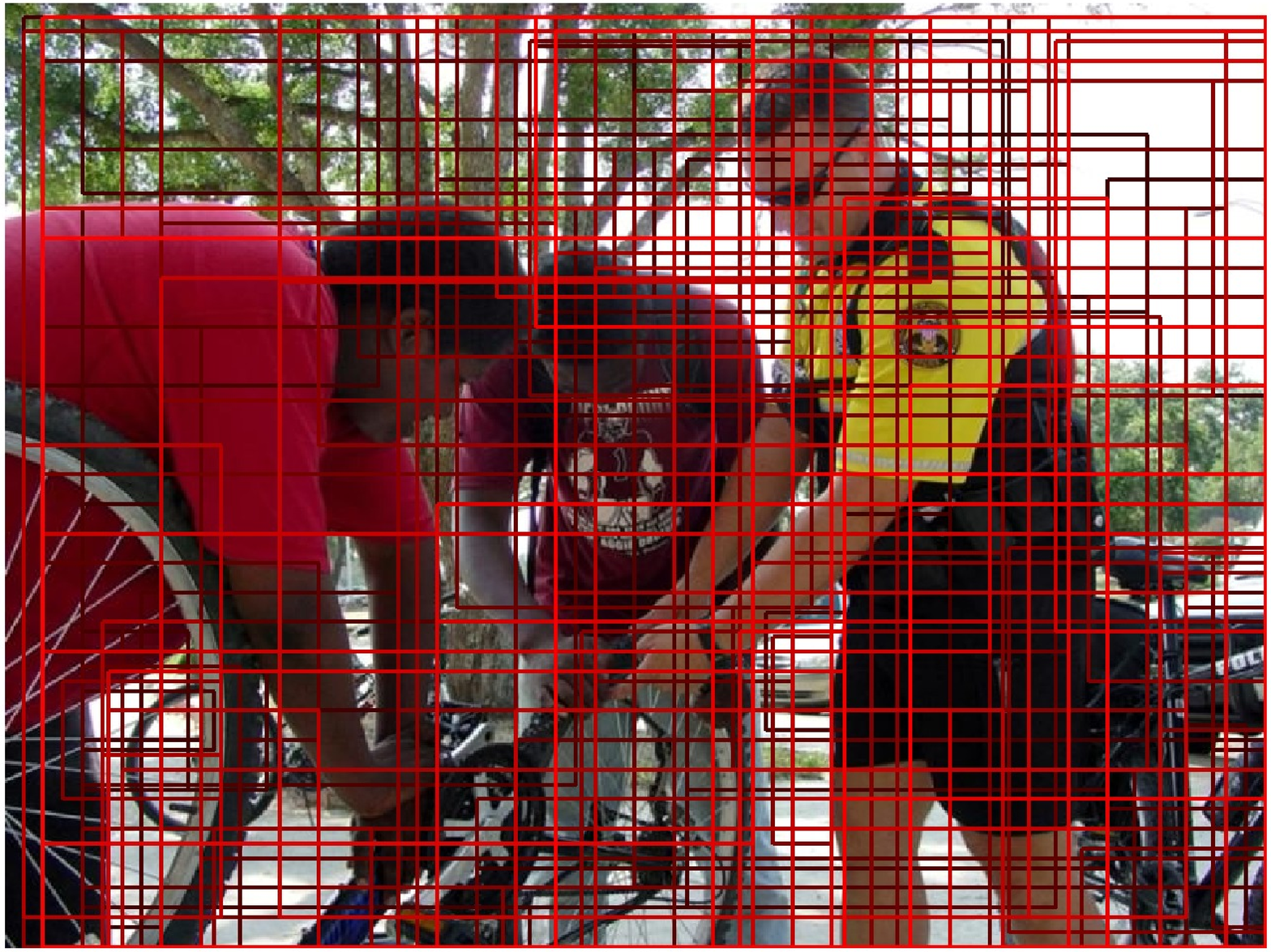}}
    \caption{Extraction region proposals from images of Stanford-40 using objectness}
\end{figure*}
\vspace{-2em}
\subsection{Performance evaluation}
In this section we evaluate the performance of our framework on action recognition over benchmark datasets with still images (Stanford-40) and videos (UCF-50). 
\subsubsection{Evaluation on Stanford-40}
We evaluate the performance of our approach in two folds. First we provide an ablation study to evaluate performance of our adaptive number of codeword selection (\S \ref{sec:code}). 
The results are shown in Table \ref{tab:t2}. In the previous work of \cite{Roy2017}, number of codewords are chosen empirically after cross validation, still adaptive selection improves that performance by $0.4\%$. 
This goes on to show that adaptive selection not only eliminate the need to execute cross-validation to select the number of code words, but also improves previous performance by a margin.

Next, we demonstrate the effect of label propagation (\S \ref{sec:lab})  in our pipeline. We compare the effect of addition of label propagation in Table \ref{tab:t22}. We achieve a performance improvement of $1.8\%$. A simple assumption that neighbouring data points tend to have similar labels clearly improves the performance of random forest. This also shows that label propagation is an effective tool to reduce the tree classifier confusions that arise in a random forest. It also adds helps majority voting, that is applied in random forest alleviate the problem of tree classifier confusions and avoid misclassification.

Finally, we evaluate the performance of our approach against other related action recognition pipelines in literature. Table \ref{tab:t1} mentions the accuracy assessment of different techniques for the Stanford-40 dataset. The performances of the methods based on hand-crafted SIFT-like features are comparatively less  ($\approx 35.2 \%$) \cite{wang2010locality}. This can be attributed to the fact that differences in human attributes for many of the action classes are subtle. Label consistent K-SVD provides $32.7\%$ accuracy. Part learning based strategies obtain better recognition performance in this respect by explicitly modelling category specific parts.
Classification accuracy of $40.7 \%$ is obtained with the generic expanded part models (EPM) of \cite{sharma2013expanded} which is further enhanced to $42.2 \%$ while the contextual information is incorporated in EPM. The best performance with shallow features obtained for this dataset is $45.7 \%$ by \cite{yao2011human} which performs action recognition by combining bases of attributes, objects and poses. Further they derive their bases by using large amount of external information.
It is worth noting that, the ImageNet pre-trained AlexNet reports a classification accuracy of $46 \%$ \cite{krizhevsky2012imagenet}. With our framework, we observe an improvement of $5.2\%$ over them. It can be argued that our  method encapsulates the advantages of deep and shallow models effectively in a single framework. The CNN based region proposals are capable of encoding high level abstractions from the local regions. Since the images are captured in unconstrained environments, the backgrounds are uncorrelated in different images of a given category. The per category dictionary learning strategy reduces the effects of such background patches and the proposed ranking measure further boosts the proposals corresponding to the shared human attributes, human-objects interaction etc. for a given action category. In contrast to other techniques which are based on SVM classifier, our framework relies on the random forest model
which does not explicitly require any cross-validation. We observe that performance of the random forest model gradually improves with growing number of CART trees within the range $500-1000$ and a random forest model with $1000$ trees outputs the best performance. Further, the addition of label propagation overcomes the problem of misclassification due to confusion created because of ensemble nature of random forest.
\begin{table*}

\centering
\caption{Effect of adaptive number of code word selection for Stanford-40}
\resizebox{\textwidth}{!}{%
\begin{tabular}{|c|c|}
\hline 
\textbf{Method} & \textbf{Classification accuracy}\\
\hline
Top $B$ (B=200 chosen empirically) codewords and LLC encoding \cite{Roy2017} & 49\% \\
\hline
\textbf{Proposed framework with adaptive number of codewords and LLC encoding} & 49.4\% \\
\hline
\end{tabular}}
\label{tab:t2}

\end{table*}

\begin{table*}

\centering
\caption{Effect of label propagation for Stanford-40}
\resizebox{\textwidth}{!}{%
\begin{tabular}{|c|c|}
\hline 
\textbf{Method} & \textbf{Classification accuracy}\\
\hline
\textbf{Proposed framework with adaptive number of codewords and LLC encoding} & 49.4\% \\
\hline
\textbf{Proposed framework with adaptive number of codewords and LLC encoding +label propagation} & \textbf{51.2\% }\\
\hline
\end{tabular}}
\label{tab:t22}
\end{table*}

\begin{table*}

\centering
\caption{A summary of the performance of our classification framework for the Stanford-40 data in comparison to the literature}
\resizebox{\textwidth}{!}{%

\begin{tabular}{|c|c|}
\hline 
\textbf{Method} & \textbf{Classification accuracy} \\ 
\hline
ObjectBank \cite{li2010object} & 32.5 \% \\ 
\hline 
label consistent K-SVD \cite{jiang2011learning} & 32.7 \% \\ 
\hline
LLC with SIFT features \cite{wang2010locality}  & 35.2 \% \\ 
\hline 
Spatial pyramid matching kernel \cite{lazebnik2006beyond} & 34.9 \%  \\ 
\hline
Expanded parts model \cite{sharma2013expanded} & 40.7 \%  \\ 
\hline
CNN AlexNet \cite{krizhevsky2012imagenet} & 46 \%  \\ 
\hline
\textbf{Proposed framework (with  adaptive number of codeword selection and LLC encoding \& label propagation)} & \textbf{51.2 \%} \\ 

\hline 
\end{tabular}}
\label{tab:t1}
\end{table*}

\begin{table*}

\centering
\caption{Effect of adaptive number of code word selection for UCF-50}
\resizebox{\textwidth}{!}{%

\begin{tabular}{|c|c|}
\hline 
\textbf{Method} & \textbf{Classification accuracy}\\
\hline
Top $B$ ($B$=200 chosen empirically) codewords \cite{Roy2017} and fisher vector encoding & 64\% \\
\hline
\textbf{Proposed framework with adaptive number of codewords and fisher vector encoding} & \textbf{64.5\%} \\
\hline
\end{tabular}}
\label{tab:t3}
\end{table*}

\begin{table*}

\centering
\caption{Effect of label propagation for UCF-50}
\resizebox{\textwidth}{!}{%

\begin{tabular}{|c|c|}
\hline 
\textbf{Method} & \textbf{Classification accuracy}\\
\hline
Proposed framework with adaptive number of codewords and fisher vector encoding & 64.5\% \\
\hline
\textbf{Proposed framework with adaptive number of codewords and fisher vector encoding + label propagation} & \textbf{66.7\%} \\
\hline
\end{tabular}}
\label{tab:t33}
\end{table*}

\begin{table*}

\centering
\caption{A summary of the performance of our classification framework for the UCF-50 data in comparison to the literature}
\resizebox{\textwidth}{!}{%

\begin{tabular}{|c|c|}
\hline 
\textbf{Method} & \textbf{Classification accuracy} \\ 
\hline 
GIST \cite{oliva2006building} & 38.8 \%  \\ 
\hline
STIP (HOG+HOF) + bag of words & 47.9 \% \\
\hline
ActionBank \cite{sadanand2012action} & 57.9 \% \\ 
\hline 
Improved dense trajectory \cite{wang2013action} & 65.3\% \\
\hline
\textbf{Proposed framework (with adaptive number of codewords and fisher vector encoding + label propagation)} & \textbf{66.7\%} \\ 
\hline 
\end{tabular}}
\label{tab:t4}
\end{table*}

\subsubsection{UCF-50 dataset}

For UCF-50, we divide our experiments in similar way we do for Stanford-40. Table \ref{tab:t3} shows the effect of adaptive selection of number of codewords. Similar to that of Standford-40, we observe an increment of $0.5\%$ in result. Table \ref{tab:t33} shows the effect of label propagation on random forest. Label propagation improves the performance of random forest by $2.2\%$. Encoding videos properly is inherently more complex than images due to added difficulty of encapsulating changes along progression of time. This in turn creates confusion in an ensemble setting such as random forest. Label propagation works well in this scenarios which is evident from such a high increment of result.  

We compare the performance of our framework with that of three different shallow representations from the literatures with similar train-test split (Table \ref{tab:t4}). 
The standard STIP (HOG + HOF) with the BoW encoding and the frame based GIST descriptors \cite{oliva2006building} exhibit  classification performances of $47.9\%$
and $38.8 \%$ respectively. Since the differences between many of the action classes in UCF-50 are fine-grained and the videos contain substantial camera motion and cluttered
backgrounds, models based on global descriptors fails drastically in discriminating the action classes. The ActionBank \cite{sadanand2012action} model based on learned 
action templates provides improved recognition performance ($57.9 \%$), although it requires numerous supervised information to learn the templates. Improved dense trajectory \cite{wang2013action}; dense trajectory with additional RootSIFT normalization provided $65.2\%$. All the aforementioned setups are based on the SVM classifiers. 

In contrast, our framework exhibits the best average recognition accuracy of  $66.7 \%$ (we use GMM with $100$ components). The enhancement of the performance of the proposed framework is attributed to the robust ranking measure which selects recurrent and discriminative local features and reduces the effects of background patches by assigning low distinctiveness scores. This is established since the recognition performance of the system sharply decreases (recognition accuracy of $\approx 58 \%$ when all the codewords are considered) as more codewords per category are considered to build the dictionary. 
We do not compare with the methods that use deep features since deep features have greater ability to encapsulate the spatial-temporal complexities involved in a video. In this work, we do not attempt to compare abilities of different feature representations rather the effectiveness of our pipeline. 

\section{Conclusion}
\label{sec4}
We introduce a novel supervised discriminative dictionary learning strategy for the purpose of action recognition from still images as well as videos. We take advantage of the available training samples to adaptively rank local features which are both robust and discriminative. Further, we cluster the local features at the entity and category levels to eliminate the effects of features corresponding to non-recurrent or background locations. The adaptive ranking paradigm proposed in this work holds wider applications in areas including feature selection, ranked set generation for retrieval etc. The effectiveness of this dictionary learning approach is validated on challenging datasets (Stanford-40, UCF-50), on which, superior performance measures can be observed in comparison to popular techniques from the literature.

\bibliography{article}   
\bibliographystyle{spiejour}   




\end{spacing}
\end{document}